\documentclass[10pt,journal,cspaper,compsoc]{IEEEtran}

\usepackage{times}  
\usepackage{epsfig} 
\usepackage{graphicx}
\usepackage{amsmath}          
\usepackage{amssymb}         
\usepackage{rotating}
\usepackage{multirow}
\usepackage{mdframed}

\begin{document}

\title{Comment on ``Ensemble Projection for Semi-supervised Image Classification''}

\author{Xavier~Boix,
        Gemma~Roig,
        and~Luc~Van~Gool, Member, IEEE
% <-this % stops a space
%\IEEEcompsocitemizethanks{\IEEEcompsocthanksitem M. Shell is with the Department
%of Electrical and Computer Engineering, Georgia Institute of Technology, Atlanta,
%GA, 30332.\protect\\
%E-mail: see http://www.michaelshell.org/contact.html
%\IEEEcompsocthanksitem J. Doe and J. Doe are with Anonymous University.}% <-this % stops a space
%\thanks{}
}

\markboth{Comment on ``Ensemble Projection for Semi-supervised Image Classification''}%,~Vol.~6, No.~1, January~2007
{Shell \MakeLowercase{\textit{et al.}}: Bare Demo of IEEEtran.cls for Computer Society Journals}

\IEEEcompsoctitleabstractindextext{%
\begin{abstract}
In a series of papers by Dai and colleagues~\cite{EnPar,EnPro}, a feature map (or kernel) was introduced for semi- and unsupervised learning. This feature map is build from the output of an ensemble of classifiers trained without using the ground-truth class labels. In this critique, we analyze the latest version of this series of papers, which is called Ensemble Projections~\cite{EnPro}. We show that the results reported in~\cite{EnPro} were not well conducted, and that Ensemble Projections performs poorly for semi-supervised learning.
\end{abstract}

% \begin{keywords}
% Ensemble Projections, Ensemble Partitioning.
% \end{keywords}
}

\newcommand{\comment}[1]{}

\def\bigO2{\mbox{${\cal O}$}}
\def\bigO{O}

\def\mA{\mathcal{A}}
\def\mB{\mathcal{B}}
\def\mC{\mathcal{C}}
\def\mG{\mathcal{G}}
\def\mV{\mathcal{V}}
\def\mE{\mathcal{E}}
\def\mF{\mathcal{F}}
\def\mH{\mathcal{H}}
\def\mL{\mathcal{L}}
\def\mN{\mathcal{N}}
\def\mS{\mathcal{S}}
\def\mT{\mathcal{T}}
\def\mW{\mathcal{W}}
\def\mX{\mathcal{X}}
\def\mY{\mathcal{Y}}
\def\1n{\mathbf{1}_n}
\def\0{\mathbf{0}}
\def\1{\mathbf{1}}

\def\balpha{\mbox{\boldmath $\alpha$}}
\def\bdelta{\mbox{\boldmath $\delta$}}
\def\bzeta{\mbox{\boldmath $\zeta$}}
\def\bphi{\mbox{\boldmath $\phi$}}
\def\btau{\mbox{\boldmath $\tau$}}
\def\bmu{\mbox{\boldmath $\mu$}}
\def\bsigma{\mbox{\boldmath $\sigma$}}
\def\bSigma{{\bm \Sigma} }
\def\btheta{\mbox{\boldmath $\theta$}}
\def\dbphi{\dot{\mbox{\boldmath $\phi$}}}
\def\dbtau{\dot{\mbox{\boldmath $\tau$}}}
\def\dbtheta{\dot{\mbox{\boldmath $\theta$}}}
\def\bGamma{\mbox{\boldmath $\Gamma$}}
\def\bDelta{\mbox{\boldmath $\Delta$}}
\def\blambda{\mbox{\boldmath $\lambda $}}
\def\bOmega{\mbox{\boldmath $\Omega $}}
\def\bbeta{\mbox{\boldmath $\beta $}}
\def\bupsilon{\mbox{\boldmath $\Upsilon$}}
\def\myphi{\phi}
\def\bPhi{\mbox{\boldmath{$\Phi$}}}
\def\bLambda{\mbox{\boldmath{$\Lambda$}}}
\def\bSigma{\mbox{\boldmath{$\Sigma$}}}

\def\balpha{\mbox{\boldmath{$\alpha$}}}
\def\bbeta{\mbox{\boldmath{$\beta$}}}
\def\bdelta{\mbox{\boldmath{$\delta$}}}
\def\bgamma{\mbox{\boldmath{$\gamma$}}}
\def\blambda{\mbox{\boldmath{$\lambda$}}}
\def\bsigma{\mbox{\boldmath{$\sigma$}}}
\def\btheta{\mbox{\boldmath{$\theta$}}}
\def\bomega{\mbox{\boldmath{$\omega$}}}
\def\bxi{\mbox{\boldmath{$\xi$}}}

\def\bPsi{\mbox{\boldmath $\Psi $}}
\def\bone{\mbox{\bf 1}}
\def\bzero{\mbox{\bf 0}}

\def\WB{{\bf WB}}

\def\A{{\bf A}}
\def\B{{\bf B}}
\def\C{{\bf C}}
\def\D{{\bf D}}
\def\E{{\bf E}}
\def\F{{\bf F}}
\def\G{{\bf G}}
\def\H{{\bf H}}
\def\I{{\bf I}}
\def\J{{\bf J}}
\def\K{{\bf K}}
\def\L{{\bf L}}
\def\M{{\bf M}}
\def\N{{\bf N}}
\def\O{{\bf O}}
\def\P{{\bf P}}
\def\Q{{\bf Q}}
\def\R{{\bf R}}
\def\S{{\bf S}}
\def\T{{\bf T}}
\def\U{{\bf U}}
\def\V{{\bf V}}
\def\W{{\bf W}}
\def\X{{\bf X}}
\def\Y{{\bf Y}}
\def\Z{{\bf Z}}

\def\b{{\bf b}}
\def\c{{\bf c}}
\def\d{{\bf d}}
\def\e{{\bf e}}
\def\f{{\bf f}}
\def\g{{\bf g}}
\def\h{{\bf h}}
\def\i{{\bf i}}
\def\j{{\bf j}}
\def\k{{\bf k}}
\def\l{{\bf l}}
\def\m{{\bf m}}
\def\n{{\bf n}}
\def\o{{\bf o}}
\def\p{{\bf p}}
\def\q{{\bf q}}
\def\r{{\bf r}}
\def\s{{\bf s}}
\def\t{{\bf t}}
\def\u{{\bf u}}
\def\v{{\bf v}}
\def\w{{\bf w}}
\def\x{{\bf x}}
\def\y{{\bf y}}
\def\z{{\bf z}}

\def\vbphi{\vec{\mbox{\boldmath $\phi$}}}
\def\vbtau{\vec{\mbox{\boldmath $\tau$}}}
\def\vbtheta{\vec{\mbox{\boldmath $\theta$}}}
\def\vI{\vec{\bf I}}
\def\vR{\vec{\bf R}}
\def\vV{\vec{\bf V}}

%%% Vector notation for sections 3 and 4
%%% Vector notation for sections 3 and 4
\def\mvec{\vec{m}}
\def\fvec{\vec{f}}
\def\appfvec{\vec{f}_k}
\def\avec{\vec{a}}
\def\bvec{\vec{b}}
\def\evec{\vec{e}}
\def\uvec{\vec{u}}
\def\xvec{\vec{x}}
\def\wvec{\vec{w}}
\def\gradvec{\vec{\nabla}}

\def\aM{\mbox{\bf a}_M}
\def\aS{\mbox{\bf a}_S}
\def\aO{\mbox{\bf a}_O}
\def\aL{\mbox{\bf a}_L}
\def\aP{\mbox{\bf a}_P}
\def\ai{\mbox{\bf a}_i}
\def\aj{\mbox{\bf a}_j}
\def\an{\mbox{\bf a}_n}
\def\a1{\mbox{\bf a}_1}
\def\a2{\mbox{\bf a}_2}
\def\a3{\mbox{\bf a}_3}
\def\a4{\mbox{\bf a}_4}

\def\sx{\mbox{\scriptsize\bf x}}
\def\st{\mbox{\scriptsize\bf t}}
\def\ss{\mbox{\scriptsize\bf s}}
\def\cR{{\cal R}}
\def\calD{{\cal D}}
\def\calS{{\cal S}}

\def\sigmae{\sigma}
\def\sigmam{\sigma}

\def\balpha{\mbox{\boldmath{$\alpha$}}}
\def\bbeta{\mbox{\boldmath{$\beta$}}}
\def\bdelta{\mbox{\boldmath{$\delta$}}}
\def\bgamma{\mbox{\boldmath{$\gamma$}}}
\def\blambda{\mbox{\boldmath{$\lambda$}}}
\def\bsigma{\mbox{\boldmath{$\sigma$}}}
\def\btheta{\mbox{\boldmath{$\theta$}}}
\def\bomega{\mbox{\boldmath{$\omega$}}}
\def\bxi{\mbox{\boldmath{$\xi$}}}

\newcommand{\denselist}{\itemsep -1pt}
\newcommand{\sparselist}{\itemsep 1pt}

 \newcommand{\ie}{\emph{i.e.\;}}
 \newcommand{\eg}{\emph{e.g.\;}}
 \newcommand{\etal}{\emph{et al.\;}}
 \newcommand{\etc}{\emph{etc.\;}}

\maketitle

\IEEEdisplaynotcompsoctitleabstractindextext

\IEEEpeerreviewmaketitle

%\vspace*{-1.4cm}

\section{Introduction}
In this note, we analyze the results of Ensemble Projections, which is the latest version of the method introduced by Dai and colleagues in a series of papers~\cite{EnPar,EnPro}. This method tackles the problem of semi-supervised learning by introducing a new feature map or kernel, that is learned in an unsupervised manner. We analyze Ensemble Projections~\cite{EnPro} since it is the latest paper in the series, and the code has been made public available.  

Ensemble projections builds a feature map from the output of an ensemble of classifiers trained with randomized class labels. Let $\x\in\mathbb{R}^N$ be a descriptor of an image, and let $\bphi(\x)=[\phi_1(\x),\ldots, \phi_K(\x)] \in \mathbb{R}^K$ be the feature extracted from $\x$ by Ensemble Projections. Each of the dimensions of $\bphi(\x)$, \ie $\phi_i(\x)$, is the output of a classifier, which was trained without using the ground-truth labels. The class labels are randomly generated by a procedure based on a novel exotic-inconsistency and local-consistency assumption. Generating the labels with this procedure avoids using the ground-truth labels, and hence, Ensemble Projections can be used in semi- and unsupervised tasks. For the semi-supervised task, a final classifier is computed using $\bphi(\x)$ as input features, using the ground-truth class labels of the images.  Results in~\cite{EnPro} show that Ensemble Projections outperforms previous semi-supervised methods. In the following, we show that the results were flawed, and that Ensemble Projection only improves over fully supervised linear classifiers.

\begin{figure*}[t!]
\centering
\footnotesize
\begin{tabular}{|cc@{\hspace{2em}}c@{\hspace{2em}}c@{\hspace{2em}}c@{\hspace{2em}}c|}
\hline
{\footnotesize \% of the unlabeled data used for training: } & 
\includegraphics[width =0.09\linewidth]{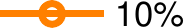}&
\includegraphics[width =0.09\linewidth]{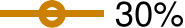}&
\includegraphics[width =0.09\linewidth]{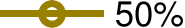}&
\includegraphics[width =0.09\linewidth]{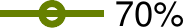}&
\includegraphics[width =0.09\linewidth]{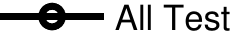}\\
\hline
\end{tabular}
\vspace{0.1cm}

\begin{tabular}{c@{\hspace{-0em}}c@{\hspace{0em}}c@{\hspace{0em}}c}
\includegraphics[width =0.25\linewidth]{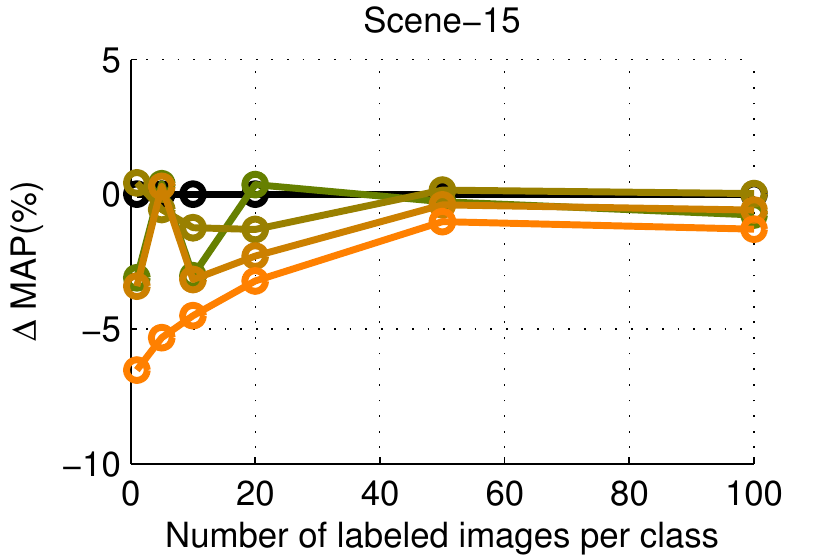}&
\includegraphics[width =0.25\linewidth]{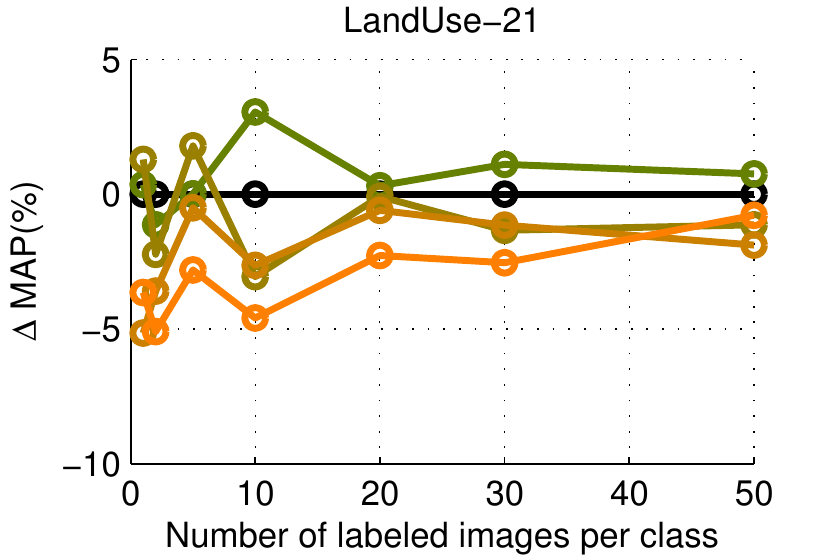}&
\includegraphics[width =0.25\linewidth]{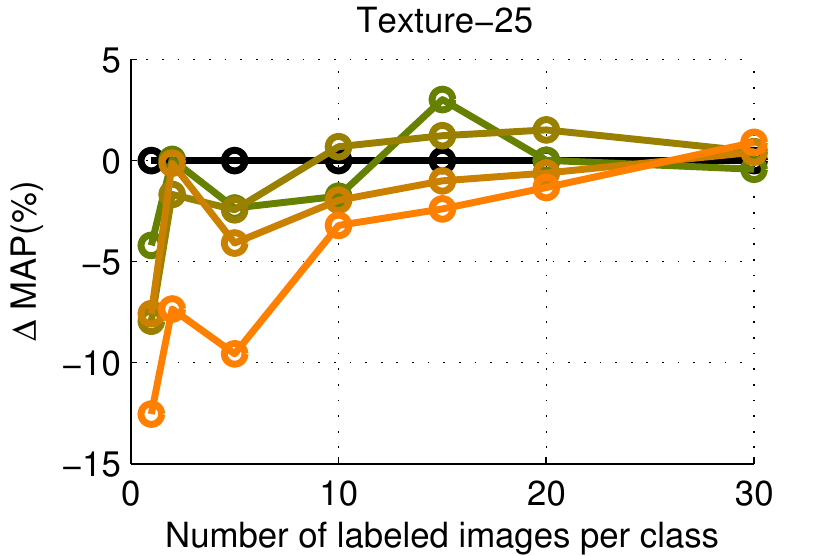}&
\includegraphics[width =0.25\linewidth]{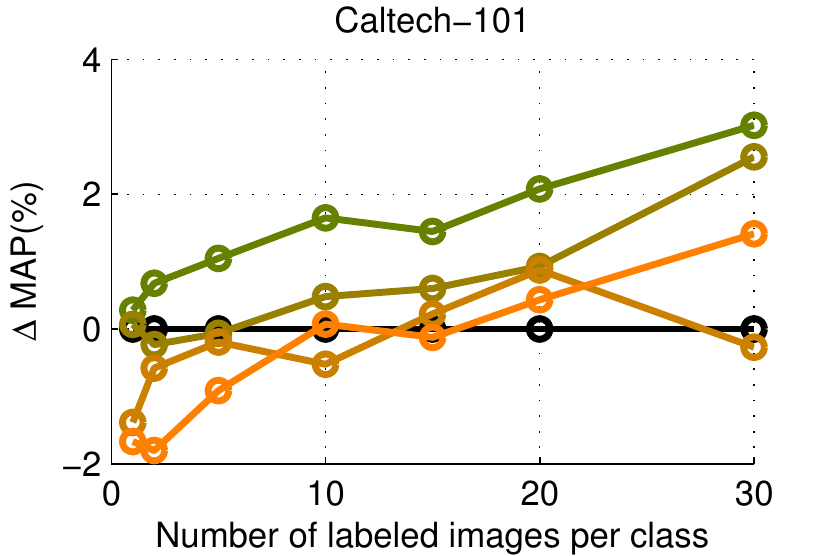}\\
 (a) & (b) & (c) & (d)  \\
\end{tabular}
\caption{\emph{Impact of using the unlabeled testing set for training.} Difference between the MAP when using the testing set for training, and for different amounts of unlabeled data (without using the testing set for training).  (a)-(d) are for the different datasets.}
\label{fig:training}
\end{figure*}

\begin{figure*}[t!]
\centering
\footnotesize
\begin{tabular}{|c@{\hspace{2em}}c@{\hspace{2em}}c@{\hspace{2em}}c@{\hspace{2em}}c|}
\hline
\includegraphics[scale =0.7]{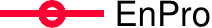}&
\includegraphics[scale =0.7]{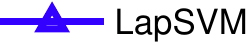}&
\includegraphics[scale =0.7]{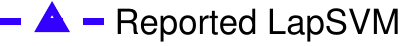}&
\includegraphics[scale =0.7]{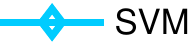}&
\includegraphics[scale =0.7]{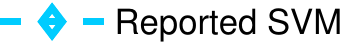}\\
&
\includegraphics[scale =0.7]{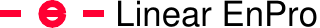}&
\includegraphics[scale =0.7]{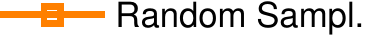}&
\includegraphics[scale =0.7]{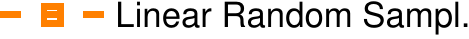} &\\
\hline
\end{tabular}
\vspace{0.1cm}

\begin{tabular}{c@{\hspace{-0em}}c@{\hspace{0em}}c@{\hspace{0em}}c}
\includegraphics[width = 0.25\linewidth]{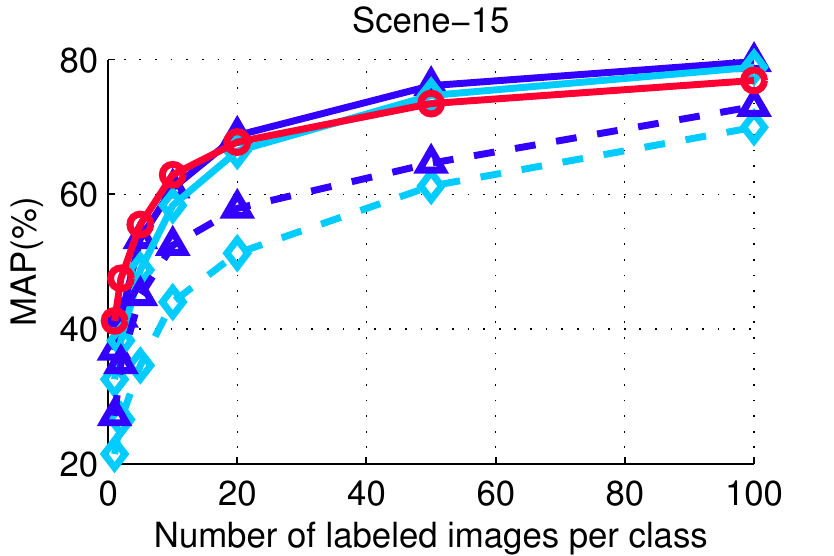}&
\includegraphics[width = 0.25\linewidth]{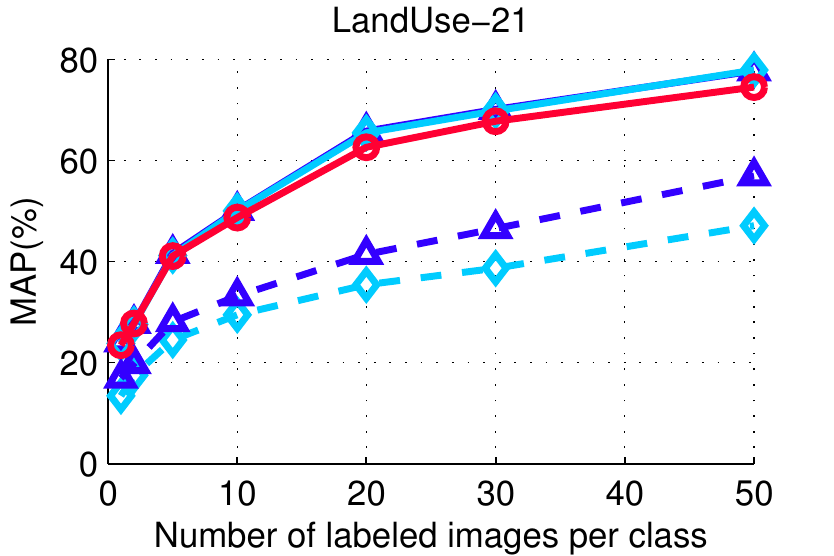}&
\includegraphics[width = 0.25\linewidth]{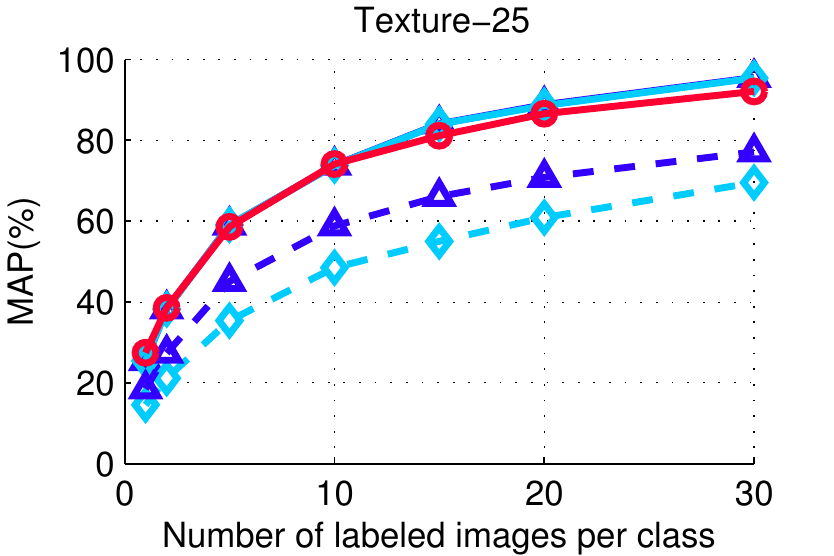}&
\includegraphics[width = 0.25\linewidth]{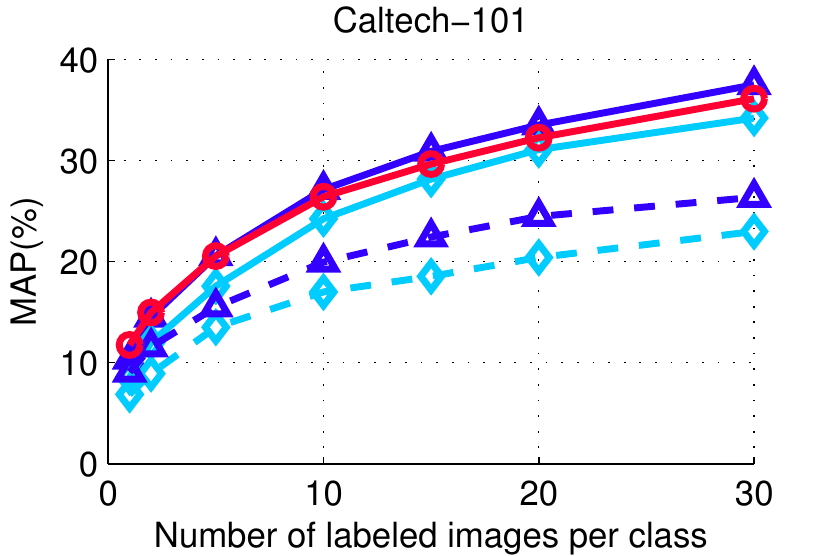}\\
\vspace{0.1cm}
 (a) & (b) & (c) & (d)\\
\includegraphics[width = 0.25\linewidth]{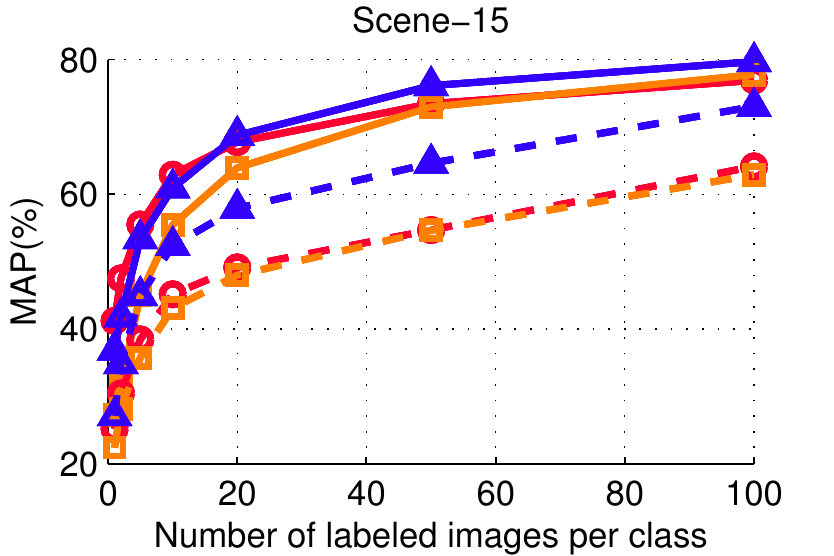}&
\includegraphics[width = 0.25\linewidth]{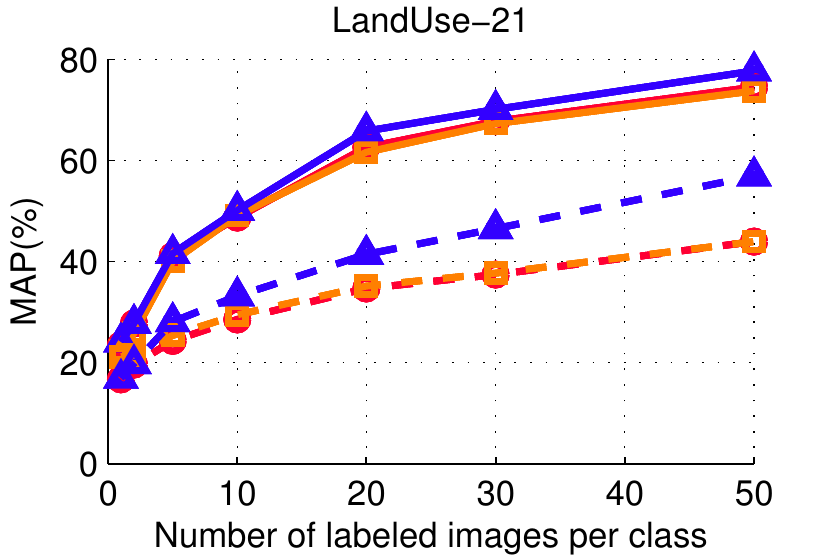}&
\includegraphics[width = 0.25\linewidth]{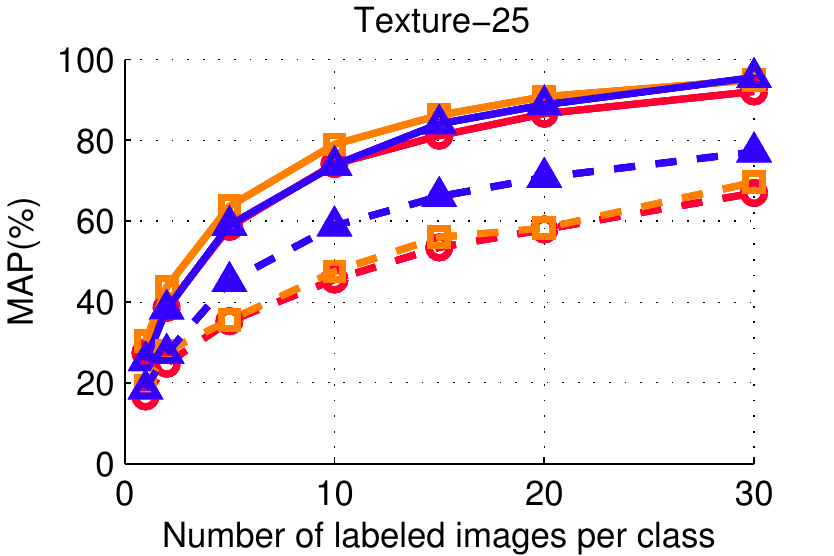}&
\includegraphics[width = 0.25\linewidth]{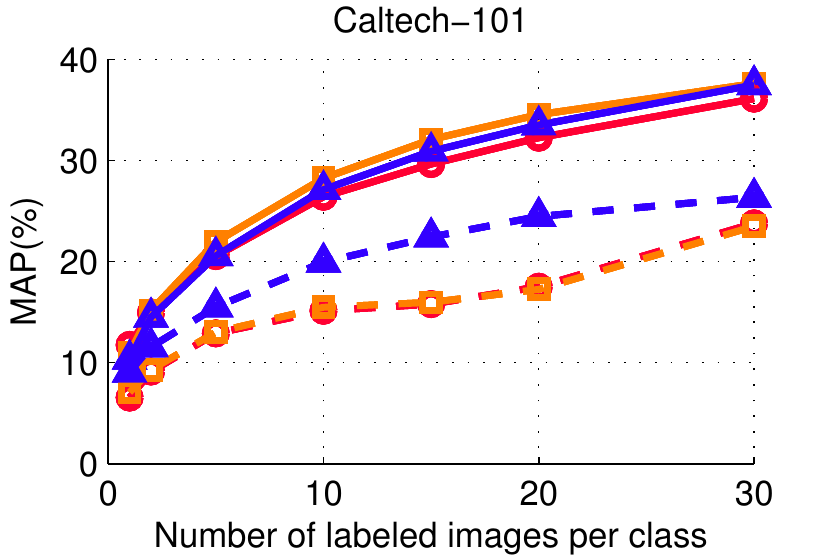}\\
 (e) & (f) & (g) & (h) \\
\end{tabular}

\caption{\emph{Comparison of EnPro with the Baselines.} (a)-(d) Comparison between the baselines reported in~\cite{EnPro} and the same baselines when using $\chi^2$ distance. (e)-(h) Evaluation of the exotic-inconsistency assumption, and the non-linearities used in Ensemble Projections. }
\label{fig:baselines}
\end{figure*}

\begin{figure*}[t!]
\centering
\footnotesize
\begin{tabular}{c@{\hspace{-0em}}c@{\hspace{0em}}c@{\hspace{0em}}c}
\includegraphics[width =.3\linewidth]{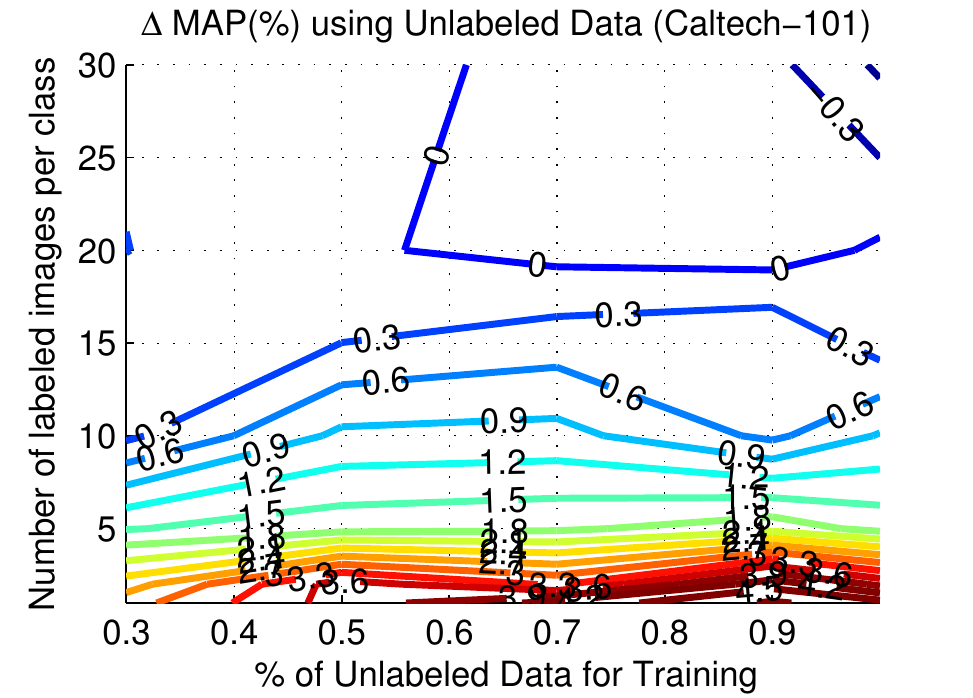}&
\includegraphics[width =.3\linewidth]{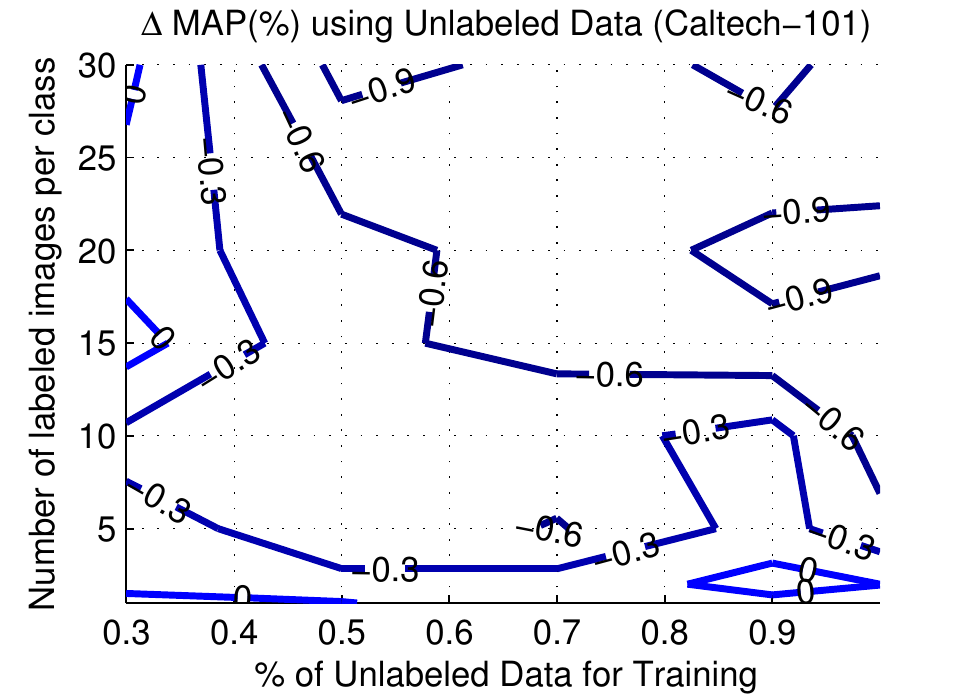}&
\includegraphics[width =.3\linewidth]{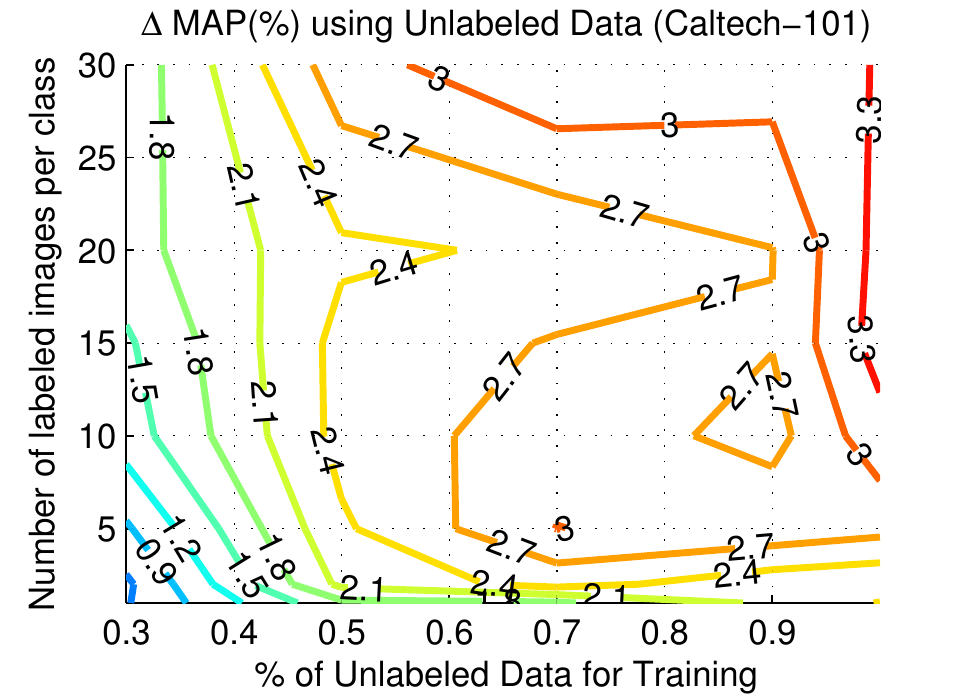}&
\includegraphics[scale =.68]{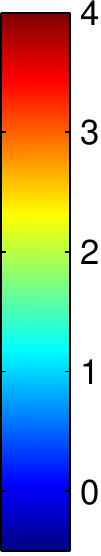}\\
(a) & (b) & (c) & (d) \\
\end{tabular}
\caption{\emph{Impact of using Unlabeled Data.} Evaluation of (a) EnPro, (b) EnPro without the exotic-inconsistency assumption, and (c) LapSVM with $\chi^2$.}
\label{fig:semi}
\end{figure*}

\section{Results}
We reproduce the results  in~\cite{EnPro} by using the publicly available code.
We use the same setup reported in the paper, 
as well as the same parameters for all methods, except when indicated. 
We use the provided datasets (Scene-15~\cite{Scene15}, LandUse-21~\cite{LandUse21}, Texture-25~\cite{Texture25} and Caltech-101~\cite{Caltech101}) as well as the same features (Gist~\cite{Gist}, LBP~\cite{ojala2002multiresolution} and PHOG~\cite{PHOG}) with the same parameters. Also, we use the same evaluation based on the Mean Average Precision (MAP), in $\%$, averaged over $5$ different random splits of the data, and report the average of the MAP. We do not report the variance since it is always lower than $2\%$ for all cases. We indicated the amount of labeled images per class, and the rest of the images are used as unlabeled data. 

In~\cite{EnPro}, results of  Ensemble Projections with both Support Vector Machine (SVM) and Support Vector Regression (SVR) were reported. We only report with SVM since the results do not significatively differ with SVR. As in~\cite{EnPro},  we use liblinear~\cite{liblinear} for the SVMs of the ensemble of classifiers to compute $\bphi(\x)$, and libsvm~\cite{libsvm} for the final classifier. 

Also as in~\cite{EnPro}, we evaluate LapSVM since it is representative of the state-of-the-art for semi-supervised learning. We use the code by~\cite{LapSVM} since it was not provided in the code by~\cite{EnPro}. Analogously to Ensemble Projections, we use the gist descriptor to  regularize the optimization problem with the unlabeled data of the training set, and we use all descriptors to learn the final classifier. Our implementation of LapSVM reproduce the results in~\cite{EnPro}.

\paragraph*{\bf{Training Using the Testing Set}}
We found that in~\cite{EnPro} the testing data was used as unlabeled data for training. The testing data should not be used for training, even if the class labels are not used, since having access to the distribution of the testing data may be (artificially) advantageous. In Fig.~\ref{fig:training}, we show the effect of removing the testing data from the learning. We report the difference of MAP between including the testing data for training, and using different amounts of unlabeled data excluding the testing set.
We report the amount of unlabeled data used for training as the percentage used over all unlabeled data. For testing, we use the unlabeled data not used for training. In Fig.~\ref{fig:training}, we can observe that using unlabeled testing data for training is advantageous, specially when few labeled images are used for training.

In the rest of the experiments in this paper, we do not use testing data for training.  We use $50\%$ of the unlabeled data for training, and the rest of data for testing.   

\paragraph*{\bf{Corrections of the Baselines}}
In the previous version of Ensemble Projections,~\ie\cite{EnPar}, the same descriptors are used for evaluation (Gist~\cite{Gist}, LBP~\cite{ojala2002multiresolution} and PHOG~\cite{PHOG}). These descriptors are accompanied with  $\chi^2$ distance, because ``\emph{the $\chi^2$ distance measure is superior for histogram-based features}"~\cite{EnPar}. Yet, in \cite{EnPro}, a linear kernel was used instead of $\chi^2$. In the first row of Fig.~\ref{fig:baselines}, we compare the results of the reported baselines when using the $\chi^2$ kernel. The $\chi^2$ is applied independently for each feature and then the average distance is computed. We can see that with the $\chi^2$ distance the baselines slightly outperform Ensemble Projections. This shows that Ensemble Projections does not outperform SVM and LapSVM when these methods are correctly configured. 

\paragraph*{\bf{Validation of the Exotic-Inconsistency}}
At this point, the reader may find  quite surprising that Ensemble Projections, that apparently is a linear method, outperforms its linear counterparts. A possible reason is that the generation of the class labels with the exotic-inconsistency assumption may be beneficial. To see if this is the case, we compare the training of the ensemble of classifiers to a naive way, which is to generate $\phi_i(\x)$ by learning a classifier with uniformly randomly generated class labels. In the second row of Fig.~\ref{fig:baselines} (orange plot), we can see that results of the random sampling are similar or superior than Ensemble Projections with the exotic-inconsistency assumption. Thus, using the exotic-inconsistency assumption does not explain the performance gain over the linear counterparts. The next experiments shows a possible explanation.

\paragraph*{\bf{Not Reported Non-linearities}}
A close inspection of the code shows that the classifiers to generate $\phi_i(\x)$ are not linear projections. A flag in liblinear is activated to pass the output of the classifier through a sigmoid, to produce a probability score as the classifier output. In the second row of Fig.~\ref{fig:baselines}, the dashed line shows the effect of using simple projections without the sigmoid. We can see that without non-linearities, the performance of the Ensemble Projections drops substantially. This suggests that Ensemble Projections outperforms the linear counterparts because it uses non-linearities that improve over the linear kernel map. 

\paragraph*{\bf{Evaluation for Semi-Supervised Learning}}
Finally, we evaluate the performance gain from using extra unlabeled data in the semi-supervised learning. In Fig.~\ref{fig:semi}, we report the MAP for different amount of labeled and unlabeled data, for Ensemble Projections with and without the exotic-inconsistency assumption, and LapSVM (with $\chi^2$). Results are reported in Caltech-101, but the same was observed for the rest of the datasets. We can see in Fig.~\ref{fig:semi} that LapSVM benefits from using unlabeled data, while Ensemble Projections does not. This shows that Ensemble Projections is not usable for semi-supervised learning.

\section{Conclusions}
We improved the experimental evaluation of~\cite{EnPro}, and we found that Ensemble Projections outperforms linear SVM because it uses non-linearities, but it does not outperform the other non-linear competing methods. Also, the performance of  Ensemble Projections does not improve when using extra unlabeled data.

\bibliographystyle{IEEEtran}
\bibliography{IEEEabrv,egbib}

\end{document}